# Multi-growth stage plant recognition: a case study of Palmer amaranth (*Amaranthus palmeri*) in cotton (*Gossypium hirsutum*)

Guy R. Y. Coleman[a], Matthew Kutugata[b], Michael J. Walsh[a], Muthukumar Bagavathiannan[b]

*Guy Coleman ([guy.coleman@sydney.edu.au](mailto:guy.coleman@sydney.edu.au)) is the corresponding author*

[a]School of Life and Environmental Sciences, The University of Sydney, Brownlow Hill, NSW, Australia

[b]Department of Soil and Crop Sciences, Texas A&M University, College Station, TX, USA

## Abstract

Many advanced, image-based precision agricultural technologies for plant breeding, field crop research, and site-specific crop management hinge on the reliable detection and phenotyping of plants across highly variable morphological growth stages. Convolutional neural networks (CNNs) have shown promise for image-based plant phenotyping and weed recognition, but their ability to recognize growth stages, often with stark differences in appearance, is uncertain. *Amaranthus palmeri* (Palmer amaranth) is a particularly challenging weed plant in cotton (*Gossypium hirsutum*) production, exhibiting highly variable plant morphology both across growth stages over a growing season, as well as between plants at a given growth stage due to high genetic diversity. In this paper, we investigate eight-class growth stage recognition of *A. palmeri* in cotton as a challenging model for You Only Look Once (YOLO) architectures. We compare 26 different architecture variants from YOLO v3, v5, v6, v6 3.0, v7, and v8 on an eight-class growth stage dataset of *A. palmeri*. The highest mAP@[0.5:0.95] for recognition of all growth stage classes was 47.34% achieved by v8-X, with inter-class confusion across visually similar growth stages. With all growth stages grouped as a single class, performance increased, with a maximum mean average precision (mAP@[0.5:0.95]) of 67.05% achieved by v7-Original. Single class recall of up to 81.42% was achieved by v5-X, and precision of up to 89.72% was achieved by v8-X. Class activation maps (CAM) were used to understand model attention on the complex dataset. Fewer classes, grouped by visual or size features improved performance over the ground-truth eight-class dataset. Successful growth stage detection highlights the substantial opportunity for improving plant phenotyping and weed recognition technologies with open-source object detection architectures.

# 1 Introduction

Image-based plant analysis presents new challenges for computer vision algorithms, where the plant continuously changes morphology both within and across seasons as influenced by varying environmental conditions. This challenge is not as prevalent in the detection of morphologically static objects (e.g. cars, cups, plates), for example, in benchmark datasets such as Common Objects in Context (COCO), which are widely used as the standard for algorithm development. The reliable recognition of plants across this morphological variability is important for high throughput plant phenotyping and precision agricultural technologies such as site-specific weed control (SSWC) (Danilevicz et al., 2021; Wang et al., 2019).

For plant phenotyping, image-based plant analysis represents an opportunity to alleviate the phenotyping bottleneck, capitalizing on gains made in genomics for more effective plant breeding (Fahlgren et al., 2015; Furbank and Tester, 2011). Algorithms that are capable of detecting and quantifying variability in population and plant morphology (i.e. plant stand count, vegetative development, leaf counts, tiller counts, etc.) and plant health would assist in screening new cultivars at wider spatial and temporal scales than is currently possible with manual measurements. For field management applications, accurate plant recognition at multiple growth stages enables the adoption of more efficient and diverse management tools and options.

This paper investigates as a case study, the image-based recognition of *Amaranthus palmeri* (Palmer amaranth) in cotton (*Gossypium hirsutum*) at various growth stages. Such capabilities will enable the adoption of more efficient and diverse SSWC methods to overcome issues such as herbicide resistance, a widespread problem in *A. palmeri* (Ward et al., 2013). SSWC reduces the wastage of resources and reduces environmental risk, fostering the adoption of alternative forms of weed control (Coleman et al., 2019). Additionally, growth-stage targeted herbicidal treatments could capitalize on differing levels of sensitivity (Chahal et al., 2015). A SSWC approach, however, relies entirely on effective and reliable weed recognition (Lopez-Granados, 2011).

Although the concepts of SSWC and high throughput phenotyping have existed for decades (Coleman et al., 2022; Furbank and Tester, 2011), recent advances in computing power, algorithm performance and accessibility, as well as improvements in camera quality have witnessed a surge in image-based deep learning for more reliable use in realistic field settings (Hu et al., 2022; Jiang and Li, 2020; Wang et al., 2019). A key driver for this change has been the development of convolutional neural networks (CNN). CNNs use labeled training data to identify important image features automatically without the need to manually select defining plant and image attributes, which has typically been an impossible task for highly complex scenes. Additionally, the 'deep' or many-layered nature of these neural network algorithms lends them a much greater ability to represent more abstract and subtle plant features, improving pattern recognition performance (Bengio et al., 2013). Together, the automatic extraction and 'deep' learning of plant features improve reliability for the highly variable plant morphology and diverse environments in which plants grow (Dyrmann et al., 2016; Šulc and Matas, 2017).

The use of CNNs for plant recognition is a rapidly growing field, with much research on the recognition of many weed species in different crops, under various agricultural contexts, and with different algorithm architectures (Hasan et al., 2021; Jiang and Li, 2020). Broadly, the output from these algorithms can be divided into whole-image classification, object detection, and pixel-wise segmentation (Coleman et al., 2022). The selection of output should be determined by the required level of localization needed, accounting for the increasing annotation requirements from classification (whole-image annotations) to segmentation, where labelling occurs at the pixel level. Within the SSWC domain, there have been many attempts at the use of image classification for weed detection (Chen et al., 2021; Olsen et al., 2019; Zhuang et al., 2022); however, the lack of precise location information is a substantial drawback.

Object detection provides plant-wise location information, without the laborious task of pixel-wise labelling required for the training of segmentation models. The approach is preferable for tasks such as wheat head detection (Khaki et al., 2022) and maize tassel detection (Mirnezami et al., 2021) for varietal screening. Pixel-wise output is often unnecessary for herbicidal weed control where the use of spray nozzles provide a coarse control approach (Salazar-Gomez et al., 2021); however, segmentation-based models are required for precise weed control tools such as lasers (Rakhmatulin and Andreasen, 2020).

Numerous object detection architectures provide variable levels of performance for plant (and plant part) detection for SSWC and plant phenotyping (Hasan et al., 2021; Xu and Li, 2022). Broadly, the architectures are divided into one- and two-stage methods, the latter (e.g. Faster R-CNN) being computationally more demanding and hence slower on resource-constrained devices. One-stage approaches include algorithms such as 'You Only Look Once' (YOLO) and single-shot detectors (SSDs). Since the release of the original YOLO algorithm in 2016 (Redmon et al., 2016), the family of model architectures has grown rapidly. The models avoid the 'region proposal' stage of the two-stage architectures such as Faster R-CNN, reducing computational requirements, making them attractive for real-world applications. Within weed recognition and plant phenotyping, the YOLO series of architectures have been found to consistently outperform other object detection methods (Barnhart et al., 2022; Dang et al., 2022) and are most relevant for field use, given the greater possible frame rate on edge devices (Partel et al., 2019).

Previous attempts at the recognition of *A. palmeri* with CNNs are sparse. Barnhart et al. (2022) compared the performance of four types of object detection CNNs for *A. palmeri* recognition within a soybean crop. YOLOv5 trained on 1024 x 1024 resolution images had the highest mAP@0.5 of 0.77. A negative relationship was found between the individual effects of height and weed density on model performance; however, the influence of specific *A. palmeri* growth stages was not addressed. With a dataset of 12 weed classes in cotton and a total of 9,370 instances, Dang et al. (2022) compared the most recent YOLO derivatives evaluating 18 different architectures from YOLO v3, v4, scaled-v4, and v5. The authors found that whilst YOLOv4 models had higher accuracy than v3 or v5 (up to 95.22% mAP@0.5), the faster inference speed and strong performance of v5-nano and v5-small models suggested that they were more likely to be used for real-time, in-field use.

As the target for this case study, the high level of genetic diversity present in *A. palmeri* is a driving factor for its status as arguably 'one of the worst weeds' (Van Wychen, 2022). This diversity connotes variability in morphological and phenological traits such as appearance and germination. Thus, collecting information on growth stage may assist in three ways: (1) diagnosing the impact of plant size variability on model performance, to train more robust algorithms; (2) phenotyping individual growth stages for improved understanding of problematic growth stages; and (3) providing growth-stage specific information for targeted weed control (e.g. tailored herbicide and precision weed control treatments) that would improve resistance management.

Yet research for the identification of weed growth stages is less common. In an attempt to understand the impact of weed maturity on algorithm performance, Burks et al. (2002) presented some of the first efforts at weed 'maturity' detection using a colour co-occurrence matrix and colour indices to differentiate three growth stage classes. The model achieved up to 97% accuracy; however, only 40 images per species per growth stage were used under ideal conditions. A more recent attempt by Teimouri et al. (2018) used an image classification CNN to estimate the growth stage of 18 different weed species, including *Polygonum* spp. and *Alopecurus myosuroides* (blackgrass) from field images. The approach achieved up to 78% accuracy for *Polygonum* spp. and 46% for blackgrass based on a leaf count estimation approach. Quan et al. (2019) found that growth stage influenced the performance of Faster R-CNN and YOLOv2 object detection algorithms on maize and a generic '*weeds*' class, with increased error rate in weeds at the two- to five-leaf stage due to similarity with maize seedlings. The effect was reversed with maize seedlings, whereby the precision of detection of six- to seven-leaf maize was 2.74% lower than for all classes combined. While these works indirectly evaluated the influence of growth stage on performance, there is little research that has directly investigated how algorithms respond to the often large morphological differences between growth stages in plants. This is a topic of interest, given that cross applicability of weed recognition in different crops has been found to be difficult, suggesting issues with generalizability (Sapkota et al., 2022).

Quantifying the impact of growth parameters on model performance is particularly important for *A. palmeri*, given how rapidly this species changes morphological appearance across a season. Further, the detection of small objects (e.g. recently germinated seedlings) is a challenging problem to solve within the computer vision domain, with low resolution and often noisy instances (Li et al., 2017), likely impacting the detection of small plants disproportionately. *A. palmeri* is highly opportunistic to favourable germination conditions, resulting in the potential for a large variability in plant growth stages throughout the cropping season (Ward et al., 2013). Understanding the interactions between growth stage and model performance sheds light on the likely ability to target different weed growth stages with different weed control treatments throughout the season. Such results may also serve as a decision support tool for model and image selection when seeking to identify variable weed growth stages.

Based on previous work and the gaps in knowledge identified, the current research seeks to investigate the growth stage recognition capability of state-of-the-art object detection models and improve our understanding of growth stage and model interactions. The specific objectives of this research were to: (1) identify growth stage recognition performance for *A. palmeri* detection across the most recent YOLO series of object detection algorithms; (2) determine the influence of image size and architecture size on the detection performance for different growth stages; and (3) investigate the effect of class number, grouping, and annotation strategy on algorithm performance for optimizing growth stage recognition.

## 2 Materials and Methods

*2.1 Field Preparation*

Field experiments were conducted at the Texas A&M University Farm in Burleson County, Texas (lat. 30.537516, long. -96.422415). GlyTol-LibertyLink cotton (resistant to glyphosate and glufosinate herbicides) was sown at two planting timings on the 2$^{nd}$ and 28$^{th}$ July 2021 in an attempt to generate diverse crop-weed stage combinations. At each timing, eight cotton rows, each 1 m wide and 91.5 m long, were planted on a flat, red loam field. The field was disked before sowing to establish a clean seedbed. *A. palmeri* seed was broadcast across the field immediately prior to crop sowing and again 14 days after to generate multiple cohort structures. Infestations of other weeds such as morningglories (*Ipomoea* spp.) and Texas millet (*Urochloa texana*) in the experimental site were controlled by manual weeding and selective herbicide applications, although some surviving plants did appear in some image datasets. Irrigation was applied using an overhead irrigation system on an as-needed basis.

*2.2 Image data collection*

Images were collected using a FLIR BlackFly BFS-70S7C-C (Teledyne FLIR LLC, Wilsonville, Oregon) 7.1 MP RGB camera (3208 x 2200 pixels) from early cotton emergence through cotton canopy closure for both times of sowing. Additional *A. palmeri* images were collected from fallow areas adjacent to the cotton area. A custom data collection backpack (Figure 1) was used for image collection over cotton rows. The backpack uses an NVIDIA (Santa Clara, California) Jetson Nano embedded computer with a custom Python (Van Rossum and Drake, 2009) image collection interface for the FLIR Blackfly cameras. The interface allows the collection of field metadata (location, time of day, weather conditions) and easy visualisation/capturing of images in field conditions. The unit is designed to be wearable, portable, rugged, and adjustable for variable row widths and image collection heights. Images were collected from a target height of 1.5 m between 12:00 h and 15:00 h in variable weather conditions (full sun, overcast, patchy cloud). An X-Rite Color Checker (X-Rite, Inc., Grand Rapids, Michigan) was used to check camera settings prior to each collection.

*2.3 Image annotation*

A total of 1,228 images (5,026 total *A. palmeri* instances across all growth stages) were annotated with bounding boxes using the LabelImg (Tzutalin, 2022) image annotation tool. Images were annotated with eight different growth stage classes, based on visual appearance, to establish an eight growth stage dataset (Table 1). All image annotations were reviewed three times and corrected by a single annotator, to ensure that growth stages were consistent. Image annotation format was converted from the default PASCAL VOC provided by LabelImg into the appropriate YOLO format using a custom Python script. A single class version of the dataset was generated by combining all eight growth stage classes into a single *A. palmeri* class. This was used as a baseline to compare with multi-class training.

Towards reducing manual annotation efforts of eight classes and addressing the third aim, two approaches were compared: (1) manual grouping by visual similarity and (2) automatic grouping based on bounding box label size. For the manual approach, the eight classes were grouped based on visual similarity, resulting in *seedling*, *single-stem,* and *bushy* (plants exhibiting a spreading phenotype) classes. The plant features used by manual sorting include leaf number, color, presence of certain plant attributes (flower heads), and size. The size-based approach used a distribution of bounding box sizes to allocate each annotation into three and eight evenly distributed size classes. These two grouping strategies (plant feature-based and size-based) were used to investigate the relationship between bounding box size and manually sorted methods for separating classes. The comparison helps determine if the model was simply using size for growth stage prediction and if using fewer growth stage classes could affect performance and potentially reduce annotation complexity.

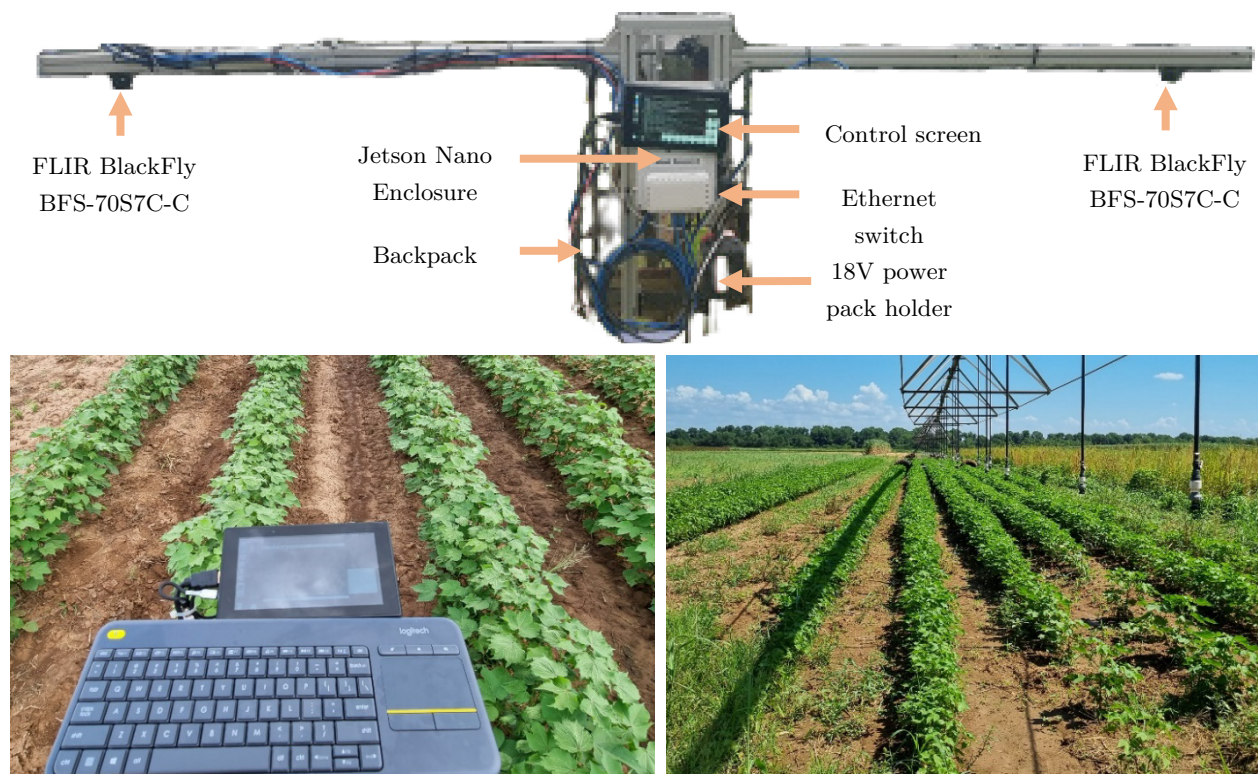

**Figure 1** The custom image collection backpack with FLIR BlackFly BFS-70S7C-C cameras (top); in field perspective on backpack interface within cotton rows (bottom left) and cotton rows under irrigator (bottom right).

*2.4 Model selection*

To capture the latest versions of YOLO models, 26 different variants from the YOLO v3, v5, v6, v6.3, v7, and v8 architectures were used for comparison (Table 2). While individual YOLO versions have been implemented previously for weed recognition purposes, and there have been recent comparisons of YOLO detectors (Barnhart et al., 2022; Dang et al., 2022), we are seeking to conduct more thorough analyses of model performance on a highly variable growth stage dataset across all recently published versions. With a substantial quantity of model variants over time, it is possible to evaluate if newer models are likely to outperform older variants or if dataset specificity is the limiting factor.

**Table 1** A total of 1,228 images (5,026 total *A. palmeri* instances across all growth stages) were annotated with eight distinct growth stages. Example instances of each growth stage are provided along with bounding box size.

| Growth Stage [†] | Description | Class ID | Instances | Bounding Box Size (cm²) [‡] | Example Instances |
|---|---|---|---|---|---|
| Seedling | ≤ 2 true leaves | PA-1 | 1052 | 3.4 ± 0.07 | 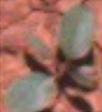 |
| 2 – 6 leaf | 3 to 6 leaves | PA-2 | 750 | 16.2 ± 0.56 | 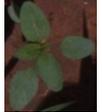 |
| >6 leaf | > 6 leaves | PA-3 | 530 | 130.5 ± 6.13 | 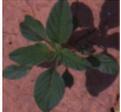 |
| Bud development | Beginning of bud development | PA-4 | 457 | 203.9 ± 16.51 | 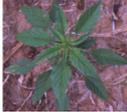 |
| Flowering | Presence of flowering inflorescence | PA-5 | 677 | 358.5 ± 20.30 | 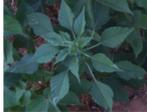 |
| Bushy – vegetative | Spreading growth habit (bushy) | SPA-1 | 460 | 401.0 ± 26.76 | 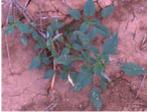 |
| Bushy – bud development | Beginning of bud development within bushy plants | SPA-2 | 514 | 859.2 ± 46.57 | 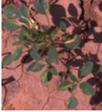 |
| Bushy – flowering | Presence of flowering inflorescence within bushy plants | SPA-3 | 586 | 1595.6 ± 68.33 | 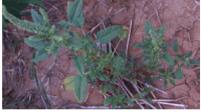 |

[†] Growth stages were decided based on visual appearance and relevance for weed control timings.

[‡] Bounding box size is presented as the mean of all instances (±SE; N=n(instances) for that class).

*2.5   Algorithm training*

All models were trained using the default, recommended hyperparameters as specified in the respective repositories (repository links are provided in Table 2). All versions of YOLO tested here included forms of default, in-built augmentation, which affects performance of the model (Dang et al., 2023). Setting hyperparameters for the training of neural network models is a lengthy process and is highly specific to the dataset on which the models are being tested. As a result, it was not attempted here, given that we are seeking to evaluate the generalizability of model performance on a complex growth stage dataset. If models were to be optimized for deployment, this hyperparameter tuning process would be required.

Models were trained and tested on Google Colaboratory Pro+ using an NVIDIA A100-SXM4-40, 40GB graphics processing unit (GPU) with PyTorch version 1.13 implementations. Towards the three aims, three different experiments were conducted: (1) test growth stage recognition performance on different YOLO versions and variants and contrast with single class performance, (2) evaluate different image and model size interactions with each growth stage, and (3) test the effect of single, three and eight classes with different annotation strategies on performance.

Across all experiments, a five-fold stratified Monte Carlo approach was adopted. Data were stratified by image collection metadata and field location. Models were trained on each fold for 30 epochs; averages with standard error (n=5) are reported. After 30 epochs, loss on the validation dataset was observed to increase, indicating overfitting, thus model training was stopped at this point. The YOLO architecture requires image dimensions as multiples of 32, thus images were resized from 3208 × 2200 and padded as needed to maintain the aspect ratio.

**Table 2** Summary of different YOLO versions and variants used. The date each repository was cloned is provided. Variant names are taken from each repository directly, and similar names do not suggest equivalent model architectures.

| **Version** | **Variant** | | | | | **Repository Link** | **Date Used** | **Reference** |
|---|---|---|---|---|---|---|---|---|
| *v3* | Tiny | Original | | SPP | | https://github.com/ultralytics/yolov3 | 16/11/22 | (Redmon and Farhadi, 2018) |
| *v5* | N | S | M | L | X | https://github.com/ultralytics/yolov5 | 16/11/22 | Github only |
| | N (P6) | | M (P6) | | X (P6) | | | |
| *v6* | N | T | S | M | L | Release 2.1 https://github.com/meituan/YOLOv6 | 16/11/22 | (Li et al., 2022) |
| *v6 3.0* | N | | L | | | Release 3.0 https://github.com/meituan/YOLOv6 | 25/01/23 | (Li et al., 2023) |
| | N (P6) | | L (P6) | | | | | |
| *v7* | Original | | | X | | https://github.com/WongKinYiu/yolov7 | 16/11/22 | (Wang et al., 2022) |
| *v8* | N | M | | X | | https://github.com/ultralytics/ultralytics | 25/01/23 | Github only |
| | | | | X (P6) | | | | |

Addressing experiment one, for growth stage recognition across YOLO versions and variants, all 26 variants (Table 2) were trained on both single and eight class datasets at image resolutions of 1280 × 896 pixels. Following the family comparison, v5 was selected for subsequent growth stage, model size, and image size analysis. The N, M, and X model sizes were trained using the eight-class dataset on 320 × 224, 640 × 448, 1280 × 896, and 1600 × 1120 pixels (image width × height). An inter-class confusion experiment was conducted on the same model and image size, where models were trained on only one of the eight classes before being tested on all eight classes to understand class confusion. Class number and annotation strategy analysis was conducted using 1280 × 896 image sizes with v5-X variant on one-, three- (manual), three- (size-based), eight- (manual), and eight- (size-based) class datasets. A total of 338 models were trained across all three experiments and data folds.

## 2.6  Class Activation Maps

A Python implementation 'pytorch-grab-cam' (https://github.com/jacobgil/pytorch-grad-cam) of Eigen class activation maps (CAMs) developed by Bany Muhammad & Yeasin (2021), was used to visualize algorithm attention during training. CAMs assist in explaining how a model is making specific predictions, by taking features from a convolutional layer in the network and assigning a weight to them based on their relative contribution to the outcome. Areas with higher red intensity indicate greater value in developing the prediction. The approach is used to uncover potential reasons why a model may be behaving in a particular manner and if other unintended aspects of an image dataset, such as the crop canopy, soil, or lighting conditions, are positively or negatively affecting predictions.

## 2.7  Performance Metrics

Model performance was compared using the common metrics of precision (i.e. proportion of total detections that are correct), recall (i.e. proportion of total possible detections that were made), and mAP@[0.5:0.95], a more thorough measure of total model performance, which takes into account performance at different overlap thresholds (i.e. intersections-over-union [IoU]) between predictions and ground truth annotations. A single IoU version, mAP0.5 was also used. Full details on the calculation of each metric can be found in Dang et al. (2022). The validation code for each YOLO version was edited to match the implementation of mAP@[0.5:0.95] in the v5 repository. When calculating performance, minimum confidence values were set to 0.001 and the IoU threshold to 0.6 for each version, the standard based on the v5 repository.

Model parameter counts and floating-point operations per second (FLOPS) were calculated with the Python '*thop*' package (https://github.com/Lyken17/pytorch-OpCounter) using image dimensions of 1280 x 1280 pixels. The Ultralytics versions v5 and v8 use an inbuilt 640 x 640 pixel resolution for FLOP calculation. This was scaled to match the 1280 x 1280 pixel resolution used for the calculation in other versions. Model efficiency was compared by calculating mAP@[0.5:0.95] per GFLOPS.

## 2.8  Statistical Analysis

Following training, each fold was considered a replicate and significant differences were assessed in R Studio (RStudio Team, 2015) using inbuilt ANOVA and Tukey's HSD tests for pairwise comparisons. All graphs were created with 'ggplot2' (Wickham, 2016). Prior to ANOVA, the normality of the data was evaluated with a residual resampling method and were considered to be normally distributed. Data transformations were not used. Bartlett's tests (P>0.05) were conducted to determine the equality of variance. Pearson's correlation analysis of on-ground pixel size by recall, and bounding box size by change in performance was conducted using the 'ggpubr' package. Correlation coefficients and significance values are provided.

## 3 Results

### 3.1 YOLO Family Comparison

In general, all YOLO versions and variants (except the smallest algorithm variants) provided mAP@[0.5:0.95] above 35% for the detection of the eight *A. palmeri* growth stages (Figure 2). The large models performed equally well (P > 0.05) with an mAP@[0.5:0.95] of up to 47.34% for v8-X. The largest variants within each version consistently performed better than the smallest variants (P < 0.05). The recently released versions (e.g. v8 and v6 3.0) did not exceptionally outperform older YOLO versions. Differences in precision and recall between the largest models was marginal, with up to 59.21% precision for v8-X variant and 64.93% recall for the v7-Original architecture (Table 3).

The mAP@[0.5:0.95] for single-class models ranged from 41.22% for v3-Tiny up to 67.05% for v7-Original variant. Single-class models, which grouped all growth stages into one class, consistently outperformed the eight class versions across all algorithms trained, by up to 16.49% and 30.51% for recall and precision, respectively (Table 3). The highest recall of 81.42% was observed for v5-X. A precision of up to 89.72% was observed with the v8-X model. The updated v6 3.0 series of algorithms did not perform as well as the v6 variants; however, the P6 versions of the same algorithms performed equally well.

Performance generally increased, moving from smaller to larger variants within each model version; however, v7-Original and v7-X versions performed comparably. The largest P6 models did not offer any significant performance benefits for this dataset. While there was no significant trend between older and newer variants, the latest v8-N variant had the highest mAP@[0.5:0.95] (P<0.05) of 58.67% on single class and 38.26% for eight-class data for all 'nano' variants for edge devices from YOLO versions v3, v5, v6, and v8.

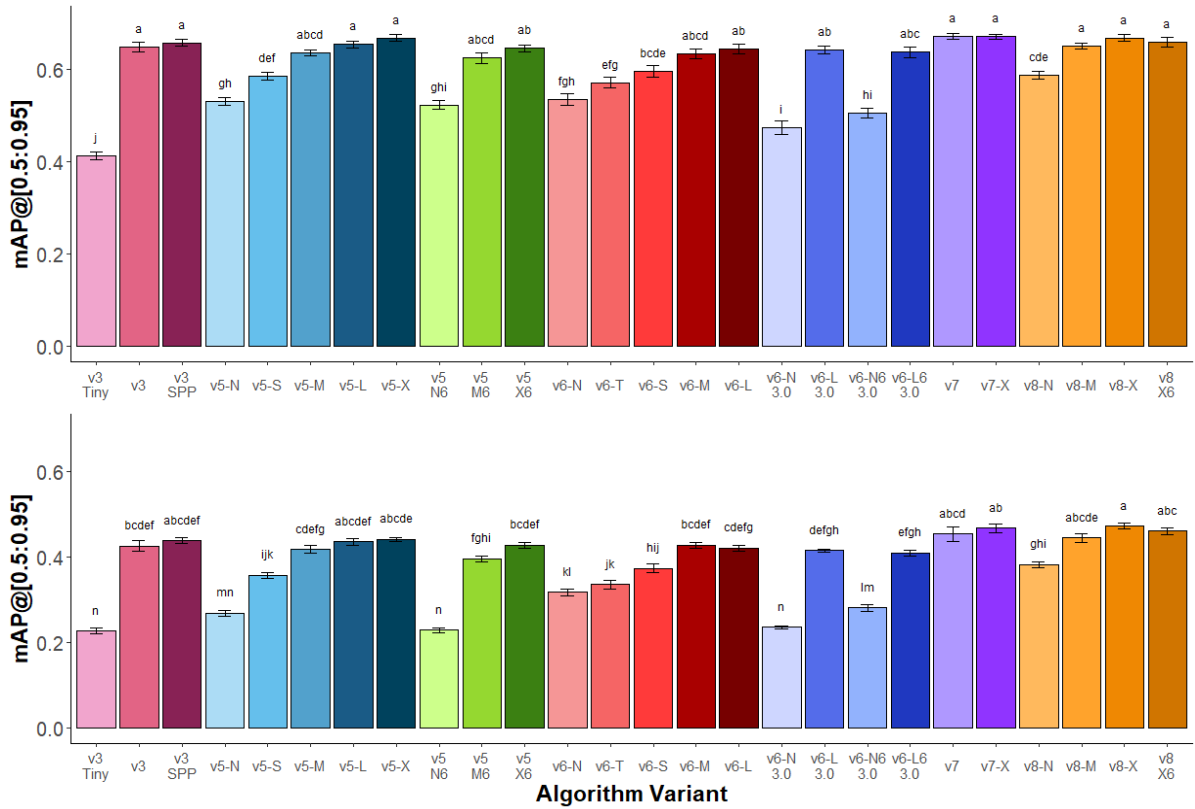

**Figure 2** Growth stage recognition performance of *Amaranthus palmeri* across YOLO versions and variants trained on a single (top) class and all eight (bottom) classes. Variants within the same YOLO version are indicated by similar colors. The largest variants within each series performed similarly. Standard error (n=5) of the mean is provided as error bars. Tukey's HSD lettering indicates significance (P < 0.05) between different variants within the single or eight class training.

**Table 3** Comparison of growth stage recognition performance of *Amaranthus palmeri* on 26 YOLO variants, including single-class performance as a baseline result. Results highlighted in red indicate the best result for that metric[†].

| Version | Variant | Precision | | | Recall | | | mAP 0.5 | | | mAP@[0.5:0.95] | | | GFLOPS | Efficiency[‡] |
|---|---|---|---|---|---|---|---|---|---|---|---|---|---|---|---|
| *Single class* | | | | | | | | | | | | | | | |
| v3 | Tiny | 80.22 | ± | 1.16 | 65.30 | ± | 0.87 | 72.81 | ± | 0.90 | 41.22 | ± | 0.90 | 52.0 | 0.79 |
| | Original | 88.43 | ± | 0.83 | 81.38 | ± | 0.59 | 87.53 | ± | 0.74 | 64.84 | ± | 1.03 | 621.2 | 0.10 |
| | SPP | 88.28 | ± | 0.89 | 80.41 | ± | 1.04 | 88.16 | ± | 0.29 | 65.66 | ± | 0.77 | 624.4 | 0.11 |
| v5 | N | 82.92 | ± | 1.29 | 69.62 | ± | 1.31 | 78.42 | ± | 0.61 | 52.89 | ± | 0.85 | **16.8** | **3.15** |
| | S | 84.40 | ± | 1.52 | 74.44 | ± | 0.93 | 82.68 | ± | 0.63 | 58.47 | ± | 0.87 | 63.6 | 0.92 |
| | M | 87.48 | ± | 0.97 | 77.86 | ± | 0.54 | 86.25 | ± | 0.46 | 63.57 | ± | 0.60 | 192.8 | 0.33 |
| | L | 87.98 | ± | 1.25 | 79.78 | ± | 1.61 | 87.53 | ± | 0.66 | 65.32 | ± | 0.71 | 432.8 | 0.15 |
| | X | 87.15 | ± | 0.83 | **81.42** | **±** | **1.11** | 88.04 | ± | 0.46 | 66.76 | ± | 0.79 | 818.4 | 0.08 |
| | N (P6) | 82.29 | ± | 1.60 | 67.85 | ± | 0.86 | 76.88 | ± | 0.80 | 52.19 | ± | 0.96 | 17.2 | 3.03 |
| | M (P6) | 85.68 | ± | 1.64 | 76.69 | ± | 0.84 | 84.34 | ± | 0.59 | 62.35 | ± | 1.08 | 197.2 | 0.32 |
| | X (P6) | 88.64 | ± | 0.96 | 78.21 | ± | 1.09 | 85.80 | ± | 0.41 | 64.48 | ± | 0.67 | 835.2 | 0.08 |
| v6 | N | 81.74 | ± | 2.53 | 62.40 | ± | 0.88 | 73.33 | ± | 1.40 | 53.39 | ± | 1.31 | 49.6 | 1.08 |
| | T | 82.17 | ± | 1.81 | 67.91 | ± | 1.11 | 77.25 | ± | 1.12 | 57.02 | ± | 1.19 | 110.6 | 0.52 |
| | S | 84.26 | ± | 0.97 | 71.09 | ± | 1.62 | 80.34 | ± | 0.96 | 59.50 | ± | 1.33 | 195.7 | 0.30 |
| | M | 86.89 | ± | 1.79 | 74.48 | ± | 1.86 | 83.85 | ± | 0.74 | 63.25 | ± | 1.07 | 360.4 | 0.18 |
| | L | 86.17 | ± | 1.99 | 75.42 | ± | 0.88 | 83.92 | ± | 0.85 | 64.33 | ± | 1.11 | 577.5 | 0.11 |
| v6 3.0 | N | 78.18 | ± | 2.75 | 58.02 | ± | 1.89 | 66.95 | ± | 1.83 | 47.30 | ± | 1.47 | 50.5 | 0.94 |
| | L | 85.08 | ± | 1.03 | 75.49 | ± | 0.58 | 83.94 | ± | 0.37 | 64.07 | ± | 0.86 | 602.0 | 0.11 |
| | N (P6) | 78.09 | ± | 1.96 | 62.10 | ± | 1.35 | 70.59 | ± | 1.17 | 50.52 | ± | 1.05 | 55.1 | 0.92 |
| | L (P6) | 86.68 | ± | 1.34 | 73.76 | ± | 1.16 | 83.79 | ± | 0.73 | 63.62 | ± | 1.16 | 675.2 | 0.09 |
| v7 | Original | 87.60 | ± | 1.15 | 81.13 | ± | 0.93 | **88.63** | **±** | **0.50** | **67.05** | **±** | **0.58** | 420.4 | 0.16 |
| | X | 87.74 | ± | 1.84 | 80.60 | ± | 1.45 | 88.27 | ± | 0.19 | 67.02 | ± | 0.55 | 755.6 | 0.09 |
| v8 | N | 85.24 | ± | 1.94 | 70.07 | ± | 0.86 | 80.01 | ± | 1.01 | 58.67 | ± | 0.91 | 32.8 | 1.79 |
| | M | 88.38 | ± | 0.68 | 76.28 | ± | 0.88 | 85.86 | ± | 0.45 | 64.97 | ± | 0.54 | 316.4 | 0.21 |
| | X | **89.72** | **±** | **0.83** | 78.06 | ± | 0.82 | 87.83 | ± | 0.31 | 66.75 | ± | 0.67 | 1032.4 | 0.06 |
| | X (P6) | 88.88 | ± | 1.29 | 77.65 | ± | 1.05 | 86.85 | ± | 0.78 | 65.86 | ± | 1.10 | 1045.6 | 0.06 |

| Version | Variant | Precision | | | Recall | | | mAP 0.5 | | | mAP@[0.5:0.95] | | | GFLOPS | Efficiency‡ |
|---|---|---|---|---|---|---|---|---|---|---|---|---|---|---|---|
| *Eight class* | | | | | | | | | | | | | | | |
| | Tiny | 40.10 | ± | 1.05 | 49.16 | ± | 1.14 | 39.91 | ± | 0.84 | 22.78 | ± | 0.60 | 52.0 | 0.44 |
| v3 | Original | 56.12 | ± | 2.21 | 62.25 | ± | 1.55 | 58.08 | ± | 1.62 | 42.68 | ± | 1.19 | 621.6 | 0.07 |
| | SPP | 57.53 | ± | 1.43 | 60.91 | ± | 1.17 | 59.55 | ± | 0.76 | 43.86 | ± | 0.66 | 624.8 | 0.07 |
| | N | 38.93 | ± | 1.70 | 52.55 | ± | 1.62 | 40.55 | ± | 1.03 | 27.00 | ± | 0.68 | **17.2** | **1.57** |
| | S | 47.33 | ± | 1.07 | 59.19 | ± | 1.50 | 50.46 | ± | 0.94 | 35.80 | ± | 0.64 | 64.0 | 0.56 |
| | M | 55.57 | ± | 1.80 | 58.44 | ± | 1.37 | 56.86 | ± | 1.02 | 41.83 | ± | 0.88 | 193.2 | 0.22 |
| v5 | L | 57.22 | ± | 1.55 | 58.64 | ± | 1.61 | 58.28 | ± | 0.81 | 43.62 | ± | 0.75 | 433.2 | 0.10 |
| | X | 57.26 | ± | 0.52 | 59.55 | ± | 1.12 | 58.60 | ± | 0.45 | 44.22 | ± | 0.47 | 819.2 | 0.05 |
| | N (P6) | 29.92 | ± | 1.15 | 53.92 | ± | 1.08 | 34.47 | ± | 0.85 | 22.97 | ± | 0.53 | **17.2** | 1.34 |
| | M (P6) | 49.25 | ± | 0.59 | 60.04 | ± | 0.61 | 53.95 | ± | 0.83 | 39.66 | ± | 0.62 | 197.6 | 0.20 |
| | X (P6) | 53.41 | ± | 0.96 | 62.23 | ± | 1.04 | 57.15 | ± | 0.85 | 42.81 | ± | 0.71 | 835.6 | 0.05 |
| v6 | N | 42.44 | ± | 1.09 | 52.19 | ± | 1.33 | 41.92 | ± | 0.95 | 31.78 | ± | 0.75 | 49.6 | 0.64 |
| | T | 44.11 | ± | 1.63 | 53.41 | ± | 2.30 | 44.33 | ± | 1.14 | 33.65 | ± | 1.01 | 110.6 | 0.30 |
| | S | 47.55 | ± | 1.83 | 55.43 | ± | 2.51 | 48.79 | ± | 1.06 | 37.41 | ± | 0.99 | 195.7 | 0.19 |
| | M | 55.61 | ± | 1.57 | 56.78 | ± | 1.37 | 55.43 | ± | 1.01 | 42.82 | ± | 0.70 | 360.4 | 0.12 |
| | L | 49.13 | ± | 1.20 | 60.57 | ± | 1.38 | 54.04 | ± | 0.99 | 42.15 | ± | 0.69 | 577.6 | 0.07 |
| | N | 45.25 | ± | 1.43 | 42.29 | ± | 2.39 | 32.13 | ± | 0.39 | 23.60 | ± | 0.33 | 50.6 | 0.47 |
| v6 3.0 | L | 53.99 | ± | 1.03 | 55.22 | ± | 0.75 | 53.01 | ± | 0.74 | 41.66 | ± | 0.34 | 602.6 | 0.07 |
| | N (P6) | 40.52 | ± | 1.35 | 49.10 | ± | 3.36 | 38.04 | ± | 0.97 | 28.23 | ± | 0.81 | 55.2 | 0.51 |
| | L (P6) | 50.34 | ± | 1.63 | 58.01 | ± | 0.95 | 52.66 | ± | 0.83 | 40.99 | ± | 0.58 | 675.3 | 0.06 |
| v7 | Original | 53.70 | ± | 2.20 | **64.93** | ± | **2.13** | 59.56 | ± | 2.33 | 45.44 | ± | 1.80 | 420.8 | 0.11 |
| | X | 59.07 | ± | 1.91 | 61.89 | ± | 1.36 | **61.14** | ± | **1.37** | 46.81 | ± | 0.98 | 756.0 | 0.06 |
| | N | 52.15 | ± | 2.55 | 55.53 | ± | 1.33 | 51.35 | ± | 1.13 | 38.26 | ± | 0.76 | 32.8 | 1.17 |
| v8 | M | 56.74 | ± | 1.79 | 58.25 | ± | 0.89 | 57.94 | ± | 0.99 | 44.54 | ± | 1.07 | 116.4 | 0.38 |
| | X | **59.21** | ± | **2.17** | 60.48 | ± | 1.56 | 60.98 | ± | 1.06 | **47.34** | ± | **0.71** | 1032.8 | 0.05 |
| | X (P6) | 54.33 | ± | 1.47 | 61.93 | ± | 1.33 | 59.30 | ± | 0.98 | 46.11 | ± | 0.83 | 1046.0 | 0.04 |

†Mean and standard error (n=5) are provided for each algorithm. Billions of floating-point operations per second (GFLOPS) are reported as an indicator of model computational requirements.

‡Efficiency is a measure of mAP@[0.5:0.95] per GLOPS.

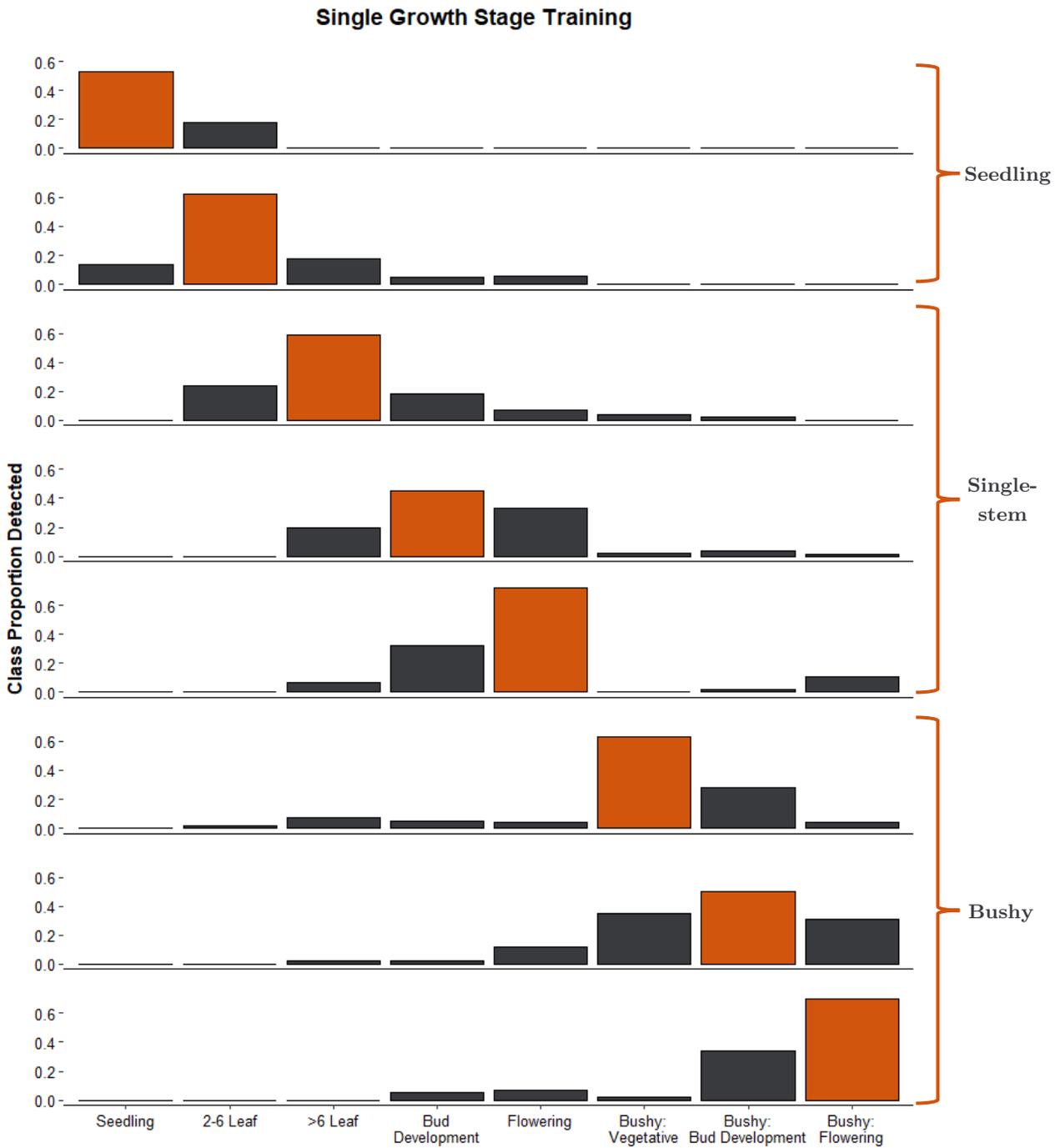

**Figure 3** Illustration of interclass misclassification between a single class trained v5-X (1280 × 896 resolution) model (orange) when tested on the full eight class dataset (grey). Class confusion was greater between more similar looking classes within the seedling, single-stem or bushy grouping.

The v5-N, v5-N (P6), and v8-N variants were the most efficient models providing efficiencies of 3.15, 3.03, and 1.79 % mAP@[0.5:0.95] per GFLOPS, respectively, on single class models. Whilst the largest models provided the best performance, they were also the least efficient. The P6 models, with an additional model layer, did not offer performance gain and were the largest models tested, offering very low-performance efficiency for edge devices, besides v5-N (P6).

*3.2 Understanding the Relationship between Growth Stage and Model Performance*

At the individual growth stage class level, the seedling stage 'PA-1' had the lowest peak mAP@[0.5:0.95] of 26.89%, on the oldest v3 SPP variant, though this was closely followed by 26.86% for v5-X and 26.85% for v8-X (Table 4). The smallest variants of each YOLO version had the lowest performance on the PA-1 and PA-2 classes (e.g., v5-N in Table 4). The seedling (PA-1) class had the smallest average bounding box size of 3.4 ± 0.07 $cm^2$. The best-performing class was SPA-3, the flowering, bushy morphology, with an mAP@[0.5:0.95] of 65.58% on variant v7-X. The flowering, single stem class, PA-5, had the highest performance of the non-bushy morphotypes at 58.10%, also with the variant v7-X.

Investigating inter-class confusion (Figure 3) confusion predominantly exists within three morphologically related groups, seedling, single-stem, and bushy. The seedling stage is only confused with two- to six-leaf stage, whilst vegetative/flowering and bushy stages are more commonly confused amongst each other. For the flowering single-stem class (PA-5), there is some confusion with bushy flowering (SPA-3); however, based on the errors for that class (Table 5), it is likely the result of misclassifying individual flowering stems that appear distinctly from the main bushy plant, though are visibly connected to the human annotator. The bud development growth stage (PA-4) was most often confused with the single-stem flowering stage (PA-5), reducing its overall mAP@[0.5:0.95]. Based on this information, these broad three-class groups (seedling, single-stem, and bushy) were used for class grouping analysis.

In collecting samples of the most confident misclassifications on the test set (Table 5), it is evident that the seedling stage is consistently confused with small grass weeds. More mature single-stem plants have confusion around the bases of bushy plants, where the vertical viewpoint makes discriminating between individual plants and bushy types more difficult.

There were misclassifications among the plants with signs of very early bud development (PA-4 and SPA-2). Further, there is confusion with the vegetative classes such as PA-3 and SPA-1. Misclassifications of crop canopy and other weeds were rare, except for morningglory (*Ipomoea* spp.) and grasses in the smallest PA-1 and PA-2 classes. For larger classes, the errors appear not to be misclassifications but rather smaller sections of the whole plant, where the IOU cutoff threshold of 0.6 is not met.

**Table 4** mAP@[0.5:0.95] performance of model versions and variants on individual classes within the eight-class trained models. Results highlighted in red indicate the best performing model for that class.[†]

| Growth Stage | Model Variant (mAP@[0.5:0.95]) | | | | | | |
|---|---|---|---|---|---|---|---|
| | v3-SPP | v5-N | v5-X | v6-L | v6-L 3.0 | v7-X | v8-X |
| **PA-1** | **26.89 ± 2.30** | 11.46 ± 1.85 | 26.86 ± 2.63 | 17.89 ± 2.20 | 19.26 ± 2.44 | 26.02 ± 4.38 | 26.85 ± 2.12 |
| **PA-2** | 36.30 ± 2.38 | 19.27 ± 1.09 | 35.57 ± 2.01 | 29.10 ± 3.32 | 31.41 ± 2.95 | **40.43 ± 1.55** | 39.53 ± 2.36 |
| **PA-3** | 43.90 ± 1.92 | 30.08 ± 3.04 | 44.45 ± 2.68 | 44.09 ± 1.96 | 46.26 ± 2.68 | 45.39 ± 2.87 | **47.12 ± 1.79** |
| **PA-4** | 36.77 ± 2.61 | 20.79 ± 2.56 | 38.18 ± 1.60 | 36.23 ± 2.58 | 30.32 ± 2.98 | 38.73 ± 2.88 | **41.54 ± 4.01** |
| **PA-5** | 52.55 ± 2.27 | 31.81 ± 3.10 | 53.91 ± 1.20 | 53.39 ± 2.21 | 52.08 ± 1.52 | **58.10 ± 2.00** | 55.33 ± 3.02 |
| **SPA-1** | 50.02 ± 1.35 | 34.45 ± 2.74 | 52.58 ± 1.78 | 52.71 ± 2.32 | 52.76 ± 1.74 | 55.65 ± 2.23 | **55.76 ± 3.63** |
| **SPA-2** | 42.52 ± 2.70 | 23.76 ± 1.61 | 40.25 ± 1.22 | 39.42 ± 1.96 | 36.84 ± 0.91 | 44.61 ± 3.79 | **47.96 ± 3.62** |
| **SPA-3** | 61.97 ± 2.23 | 44.40 ± 1.39 | 61.93 ± 2.39 | 64.39 ± 2.51 | 64.37 ± 2.17 | **65.58 ± 2.75** | 64.67 ± 2.93 |
| **Average** | 43.86 ± 0.66 | 27.00 ± 0.68 | 44.22 ± 0.47 | 42.15 ± 0.69 | 41.66 ± 0.34 | 46.81 ± 0.98 | **47.34 ± 0.71** |

[†]Standard error of the mean (SE) is provided using the five stratified Monte Carlo folds (n=5).

**Table 5** Five examples of incorrect detections with high confidence for each of the eight classes using the v5-X algorithm with an image size of 1600 x 1120.

| | | | | | |
|---|---|---|---|---|---|
| **PA-1** | 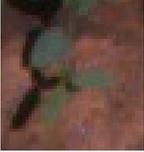 | 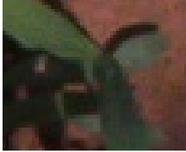 | 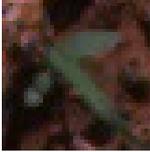 | 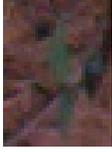 | 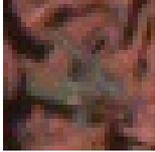 |
| **PA-2** | 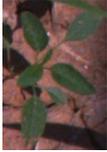 | 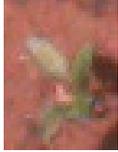 | 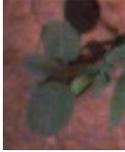 | 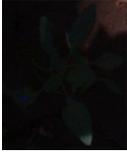 | 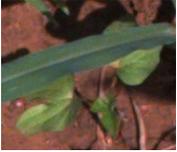 |
| **PA-3** | 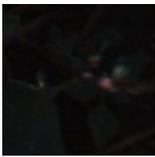 | 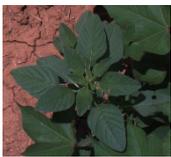 | 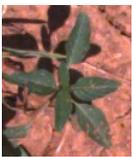 | 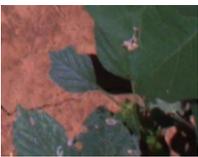 | 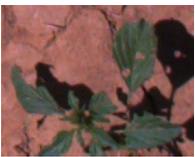 |
| **PA-4** | 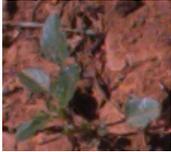 | 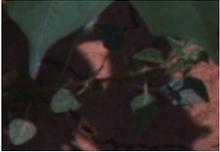 | 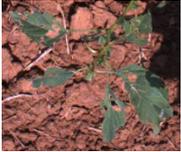 | 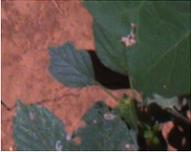 | 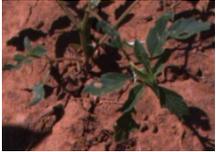 |
| **PA-5** | 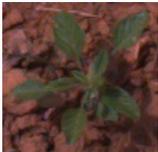 | 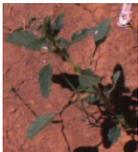 | 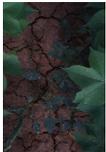 | 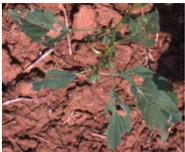 | 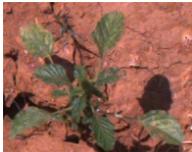 |
| **SPA-1** | 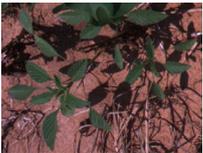 | 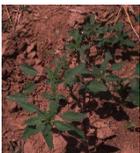 | 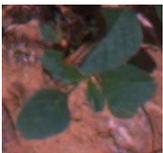 | 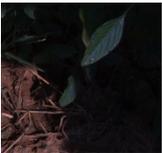 | 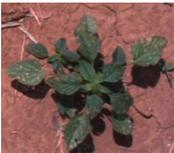 |
| **SPA-2** | 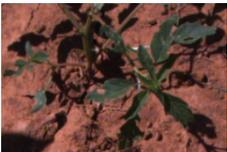 | 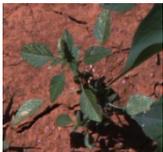 | 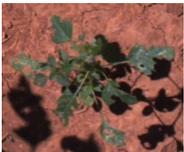 | 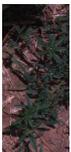 | 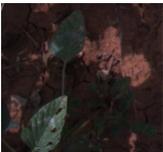 |
| **SPA-3** | 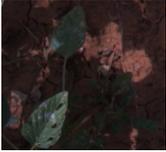 | 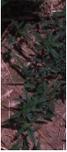 | 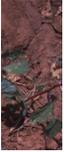 | 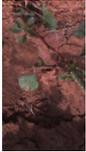 | 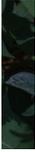 |

*3.3    Model Explainability – Class Activation Maps*

Understanding where a model focuses its attention during the training process can uncover trends hidden beneath the 'black box' of the model output (Figures 4 and 5). The presence of strong feature contributions towards predictions within the crop canopy suggests some confusion between *A. palmeri* and the cotton canopy. Attention to background conditions may indicate a reliance on some background patterns to assist predictions, indicating bias within the dataset. Some images (for example, images four and six) indicate strong attention to obvious flowering components of larger *A. palmeri* plants. Moving through the model layer-by-layer (Figure 5), the model switches attention between the canopy and the between-row conditions. Early layers identify lower-level features, such as leaves and background variability such as the remnant soil cover.

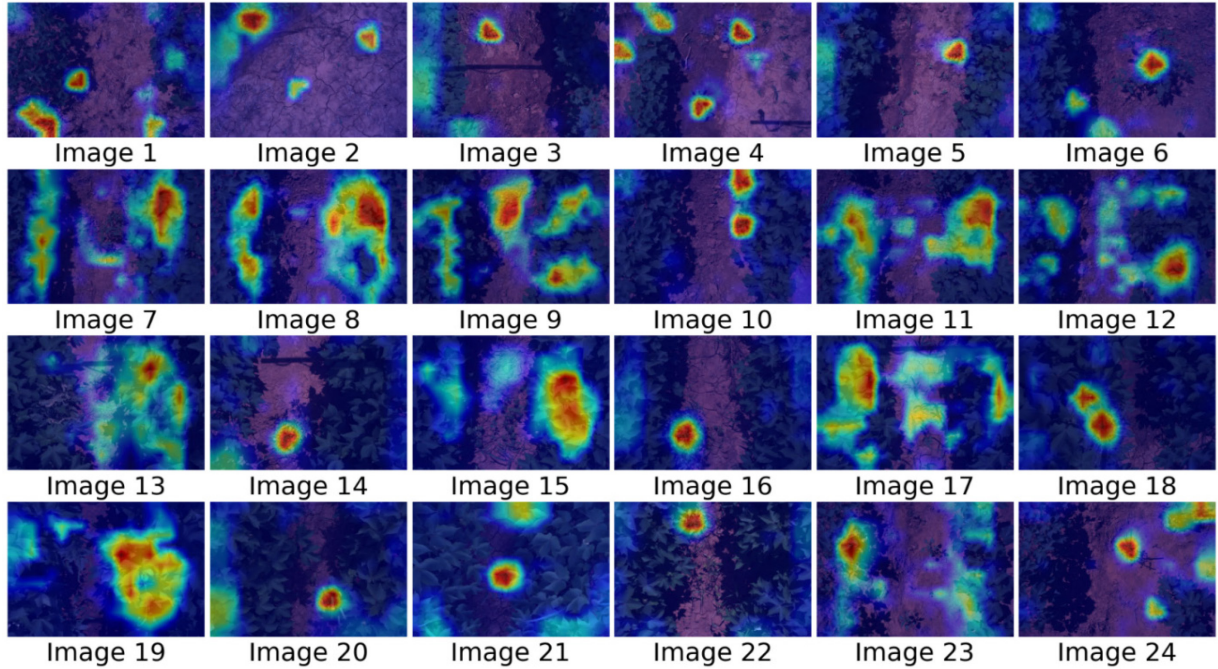

**Figure 4** A selection of test images analysed with YOLOv5 M at 640 × 448 image resolution with Eigen class activation maps (CAM) overlaid. Areas of contribution to model detection are indicated in red, whereas areas with less contribution are blue-yellow. Model confusion with the cotton canopy and some other background weeds is evident.

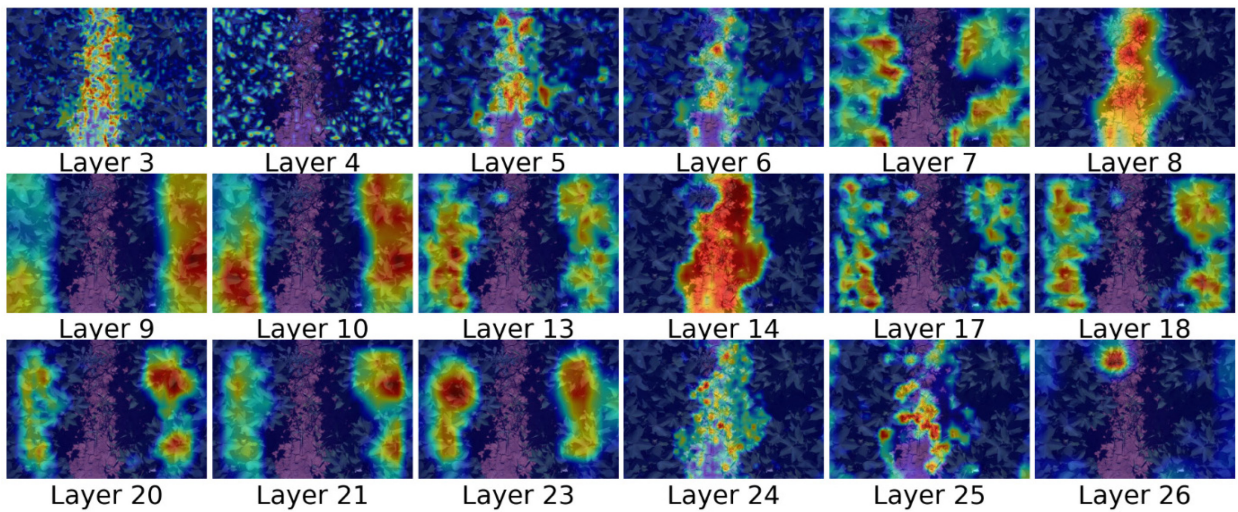

**Figure 5** Layer-wise Eigen class activation maps (CAM) through the YOLOv5 M architecture. Layers 1 through 23 correspond to convolutional layers. For efficiency, max-pooling layers are not included. The final three layers correspond to the detection head.

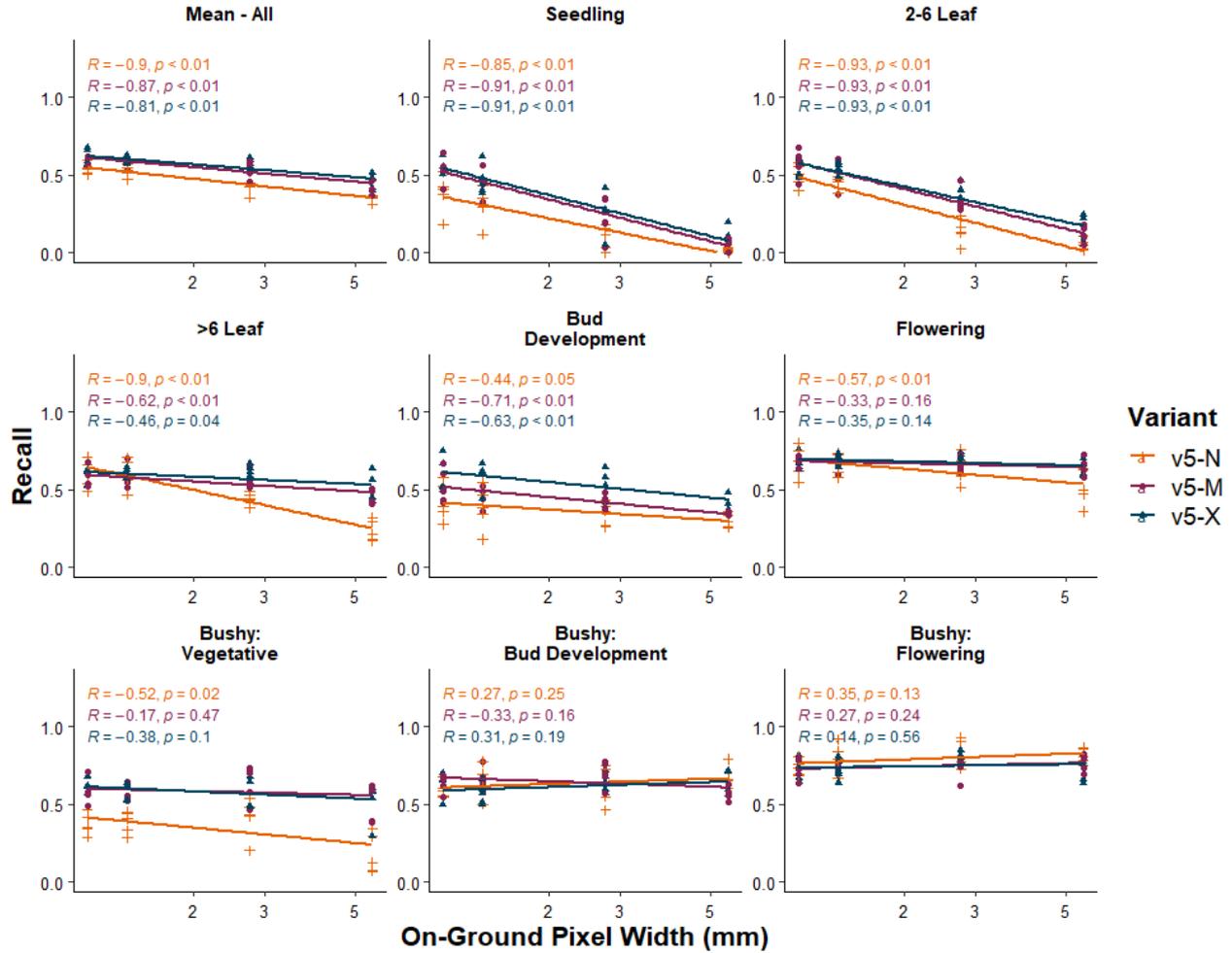

**Figure 4** Correlation analysis of recall by different on-ground pixel widths across each of the eight classes. As image resolution decreases, pixel width increases, resulting in a reduced ability to find smaller weeds. Larger model variants are more capable of managing reduced resolution; however, with the smallest weed sizes, there are still substantial reductions in performance.

*3.4  Image and Model Size Interactions with Growth Stage Detection*

Lower resolution images resulted in poorer mean recall for all v5 variants using the eight-class dataset. A decline of 15.41% for v5-X and 18.99% for v5-N was observed when reducing image resolution from $1600 \times 1120$ pixels to $320 \times 224$ pixels. Strong negative correlations were observed between on-ground pixel width and mean recall for v5-N, M and X. At the individual growth stage level, this correlation was most pronounced (P<0.05) for all models at growth stages at and below the six-leaf stage, though continuing to the Bushy Vegetative stage (SPA-1) for v5-N.

The largest drop in performance due to change in on-ground pixel size was for seedling and two- to six-leaf *A. palmeri* plants, which had the smallest mean bounding box areas of $3.4 \pm 0.07$ and $16.2 \pm 0.56$, respectively. At the seedling stage, recall for the v5-X variant dropped from 56.34% to 6.25% when image resolution decreased (increasing on ground pixel width), whilst for the *flowering* growth stage, the change was from 67.83% to 62.98% (Figure 7). At the larger than six-leaf stage, the largest variant v5-X more than doubled the performance of v5-N. Correlations were not observed (P>0.05) between on ground pixel width and recall for v5-M and v5-X for growth stages PA-5 and SPA-1 to SPA-3.

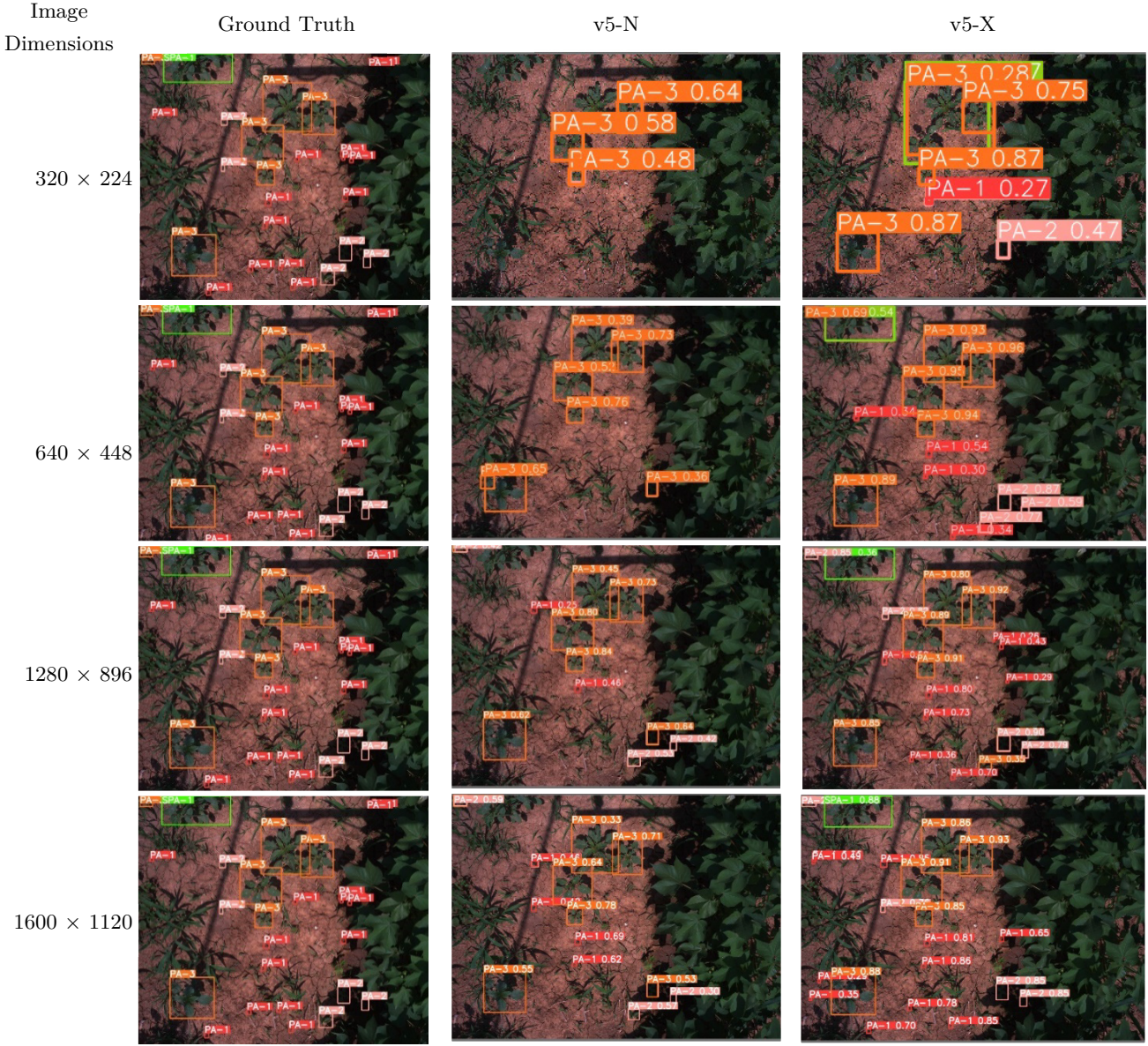

**Figure 5** Illustration of growth stage detection at the four resolutions tested for the ground truth (left), v5-X, and v5-N model variants. At lower resolutions, the difference between the smaller and larger model becomes apparent.

As with algorithm efficiency (Table 3), model size and image size directly influence the computational demand of a model. Significant negative correlations (P<0.05) were found for all metrics and variants between bounding box area and change in performance between lowest and highest resolution images (Figure 8). The result confirms that performance gain is reduced for larger objects at higher resolutions. The benefit in resolution is clearest for the smallest objects. For object sizes over 100 cm², the advantage of higher image resolution is reduced for all model sizes. For resource-constrained devices where computational power is limited, high-resolution images do not provide substantial benefit for the detection (recall) of larger objects over 300 – 400 cm²; however, high-resolution images offer substantial benefits for the smallest growth stages. Benefits to precision were retained, though highest for small objects, for all growth stage sizes, indicating a loss of image detail and increased confusion between classes. There were indications of recall and precision loss on large objects at high resolutions.

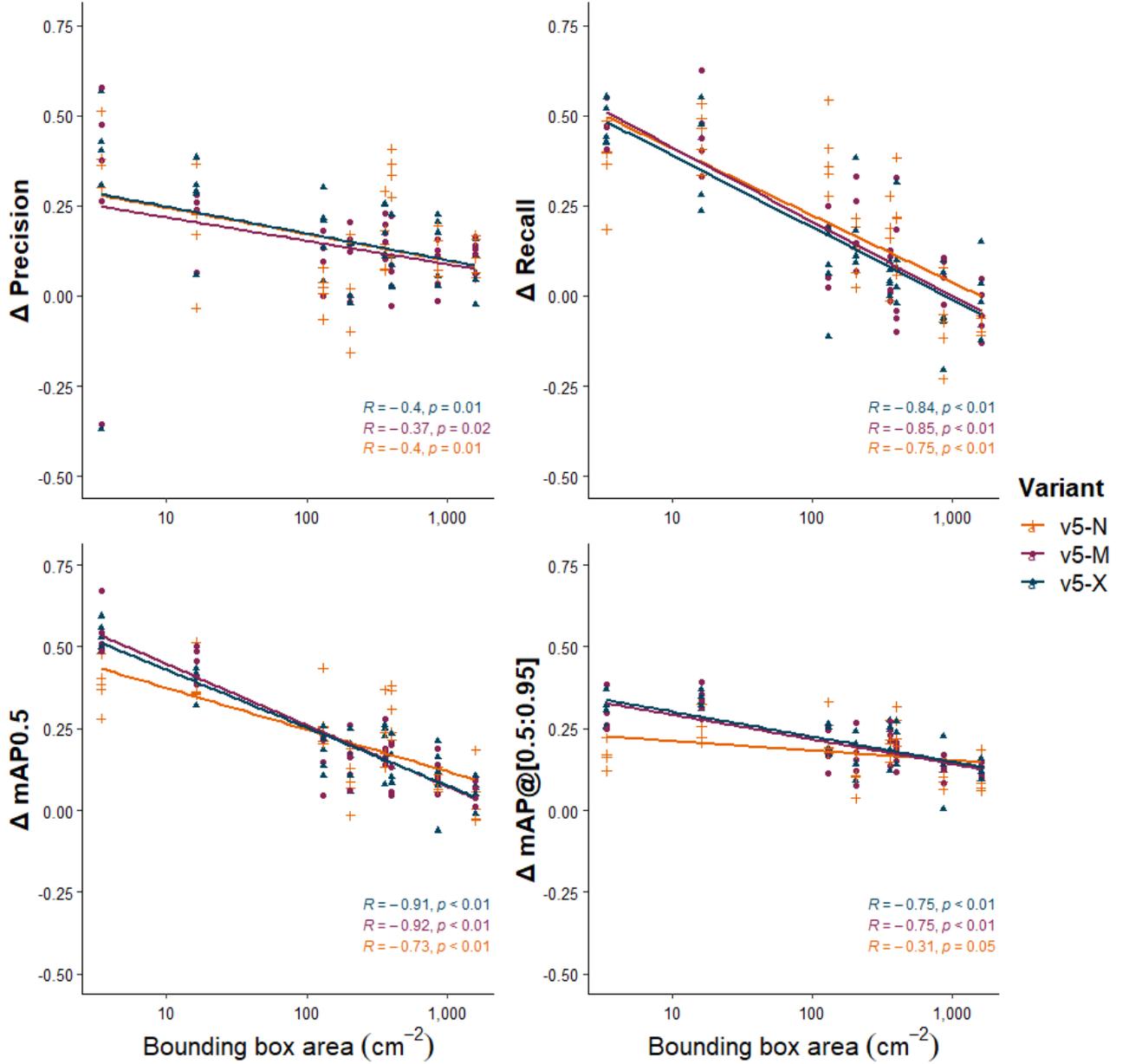

**Figure 6** Change in performance across four model metrics (precision, recall, mAP0.5, and mAP0.5:0.95) for eight-class detection by mean bounding box area for each size-based class. Change in performance was calculated by comparing performance between 1600 × 1120 and 320 × 224 resolution images. As bounding box area increases, the improvement in performance is reduced. Change in recall is the most affected by increasing bounding box area, suggesting that small plants are often missed at low resolutions.

3.5  *Interaction between Class Number and Annotation Method on Performance*

Reducing the number of classes from eight, to three, to one significantly improved (P<0.05) performance on manually annotated datasets for all metrics using the v5-X model at a resolution of 1280 × 896 pixels (Figure 9). The biggest change was observed for precision, which increased from 57.26% to 87.15%. Changes in precision reflect the reduced possibility of class confusion, whereas for single-class models, reduced precision is only caused by false detections. For eight-class models, this reduced precision occurs with inter-class confusion (Figure 3).

Whilst there was no change (P>0.05) in performance between size-based and manual grouping for three-class models, size-based grouping for eight-class models significantly improved performance. This improvement (P<0.05) was observed in the largest three classes (Figure 10). The smallest class grouping performed worse (P<0.05) than the manually defined seedling growth stage class. In general, performance declined as the bounding box area of each class declined. This pattern was not observed for manual groupings, where the performance of *bud development* classes was worse than the adjacent *vegetative* or *flowering* classes.

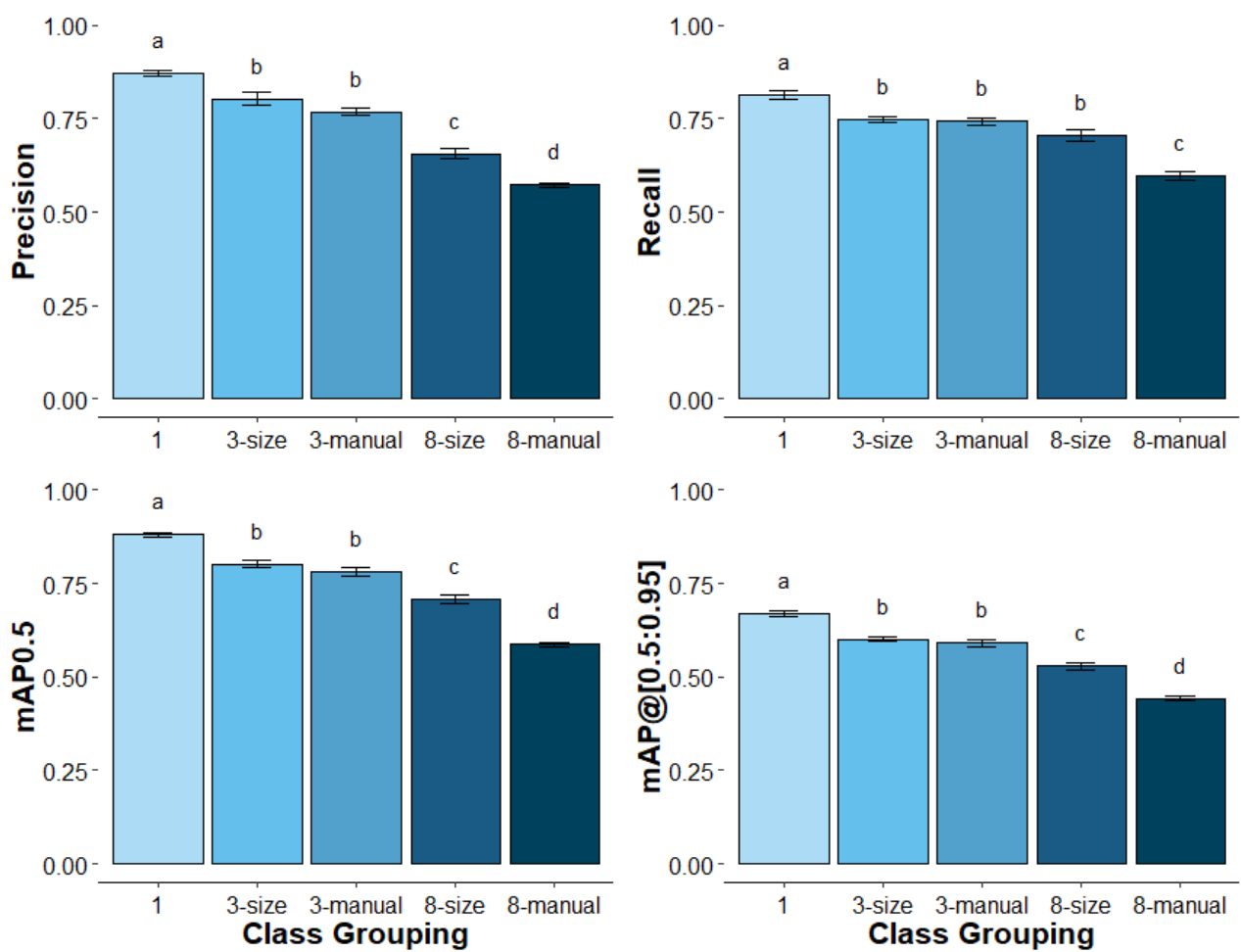

**Figure 7** Change in performance with different class grouping strategies for v5 X models at 1280 × 896. More classes reduced performance in all metrics. Grouping manually using expert-identified visual features or by automatically by bounding box size did not change performance significantly (p < 0.05).

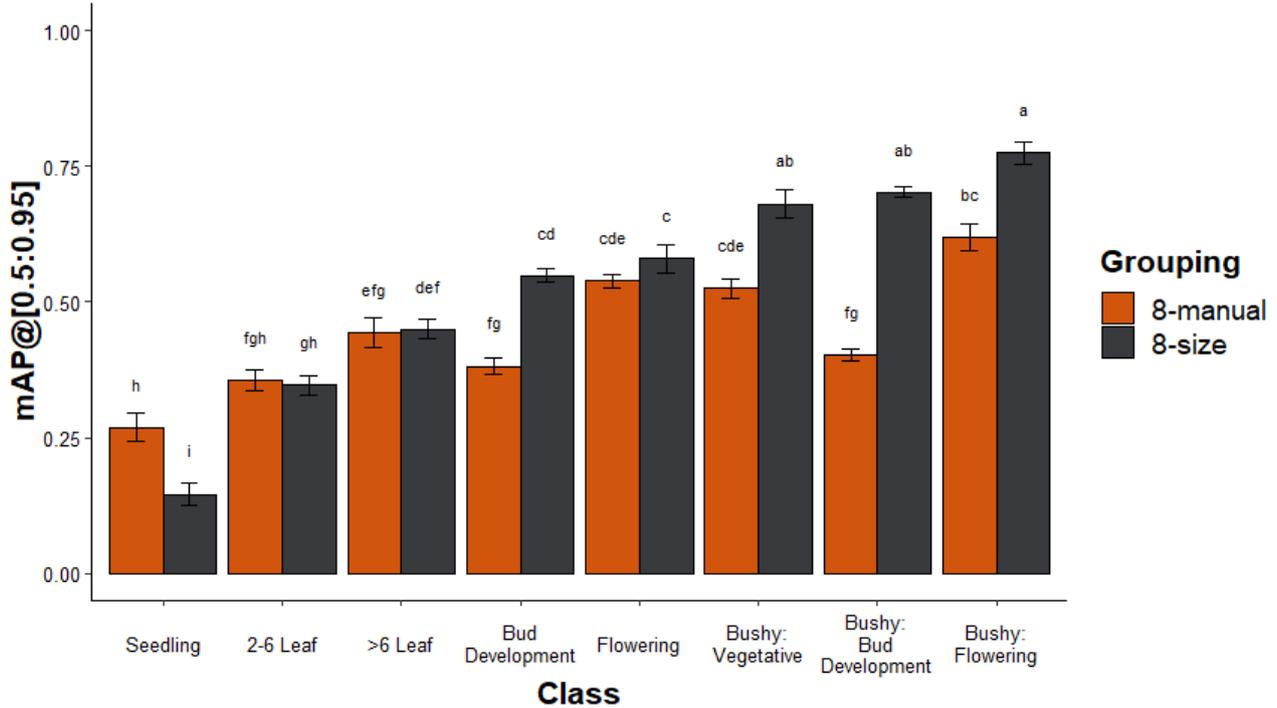

**Figure 8** Comparison of class-wise performance for manually grouped (orange) and size grouped eight-class (dark grey) YOLOv5 X models trained at 1280 × 896 image resolution. Manual class labels are provided for comparison at each class, though do not reflect the size-based groups, which were calculated automatically for an even distribution of instances in each size-grouping. Error bars are provided with standard error (N=n(instances)) of the mean. Tukey's HSD lettering indicates significance (P > 0.05) between different variants between class groups.

## 4 Discussion

Growth stage recognition is an important tool for both plant phenotyping and weed recognition purposes, with this research indicating that large YOLO variants are capable of differentiating across all eight growth stages of *A. palmeri* considered here. More recent YOLO versions did not offer significant improvements over older versions; however, larger variants within each version offered substantial increases in precision, recall, and mAP@[0.5:0.95]. These results demonstrate that growth stage recognition is possible with relatively easy-to-use object detection architectures for plant science and precision agricultural applications.

Understanding the response of popular and high-performing algorithms to the variability of growth stages is an important step in improving the high throughput field phenotyping for plant science and precision agricultural fields such as SSWC. Despite the potential improvements in phenotyping and increasingly targeted weed control offered by growth stage recognition, research on growth stage analysis has been sparse and largely limited to regression models (Barnhart et al., 2022; Burks et al., 2002) rather than the targeted detection of individual plants at certain growth stages (Teimouri et al., 2018). This work shows that growth stages interact differently with object detection models.

### 4.1 Performance of YOLO Architectures on Multiclass Plant Datasets

In the highly variable, realistic field conditions represented by the collected dataset (variable lighting conditions, soil backgrounds, and crop-weed growth stages), v5-X was capable of recalling up to 81.42% of *A. palmeri plants* with a precision of 87.15% and mAP@[0.5:0.95] of 66.76% for single class detection. These results are lower than that of Dang et al. (2023), who achieved an mAP@[0.5:0.95] of up to 88.70% with the same v5-X architecture on a 12-class dataset of different weeds. The images in Dang et al. (2023) are relatively constrained to simple soil backgrounds, similarly sized plants within a class, and distinct classes with clear visual differences. Barnhart et al. (2022) attempted the detection of *A. palmeri* in soybean and achieved a peak mAP0.5 of 77% for v5-X, compared to an mAP0.5 of 88.04% achieved here. The performance on a dataset of diverse growth stages is promising; however, the substantial decline in performance for eight-class models highlights the challenge of the task. The mAP0.5 for v5-X dropped to 58.60%, while the peak performance for eight classes achieved by variant v7 of 61.14%, well below the results achieved by both Barnhart et al. (2022) and Dang et al. (2023).

Though desktop-based comparisons provide valuable insights into model performance, the real test of algorithm performance, generalizability and applicability in real-world situations occurs in field use on edge devices. The metrics provided here do not necessarily convey the likely in-field performance, as suggested by Salazar-Gomez et al. (2021). Fortunately, the improvement in performance efficiency with newer versions of YOLO suggests that the 'N' and 'S' variants will provide increasingly high performance for real-time growth stage recognition. Operating an 'X' model on 1280 × 1280 resolution is unlikely for resource-constrained devices such as the Raspberry Pi (combined with external processing) or the lower-cost NVIDIA Jetson series of devices. For larger weeds, image resolution is not so critical; however, the resolution appears to offer greater benefits than larger model architectures. The v8-N and v5 N offer promising levels of performance with model sizes that are realistic for field deployment on precision sprayers or field phenotyping equipment.

*4.2 Growth stage, class grouping, model, and image size*

A key finding is the increased performance from using a single class for all growth stages, instead of breaking data into all eight or three classes, irrespective of grouping strategy. This is in contrast to other research that found improved performance when using two classes for tea shoot detection than a single class (Li et al., 2021), indicating the potential for species- or class-specific responses. The suggestion in that research was that by dividing data into multiple classes, the within-class variability was reduced, and the model was able to learn patterns more effectively. In the case of the continuity of growth stages presented here, the visual differences between classes were less clear than the tea shoots. Assigning discrete, individual classes to a continuous dataset likely results in increased inter-class confusion, (Figure 3). This is evident in the 30% drop in peak precision between eight- and single-class datasets, compared to a drop in recall of 16%. The difference between a drop in precision and recall suggests the overall decline in performance is largely the result of similar classes being confused, rather than missed weeds. For field use in spot-applying herbicides where weed misses are most critical, this is less of an issue.

Similar to Quan et al. (2019), the lowest performance was in seedling-stage plants; however, the result is likely caused by image resolution. Image resolution impacted the detection of the smallest weeds most and was not recoverable with larger variants, though they did provide some benefit (Figure 6). As image resolution increases, on-ground pixel width decreases, providing more pixels per unit area, which is particularly important for objects represented by very few pixels. Small object detection is a known challenge for object detection algorithms (Li et al., 2017). Resolving this issue would require reducing camera height or partitioning large images into smaller sections, given that training and deployment on full-resolution images is not currently possible due to memory and computational limitations. Importantly, in high-resolution images the detection of both small and large objects in the same trained algorithm appeared effective, indicating that modern algorithms can manage diverse appearances of target classes.

These growth stage results challenge the suggestion by Barnhart et al. (2022) of a linear and predictable relationship between the growth stage and model detection performance. Given the purposeful non-linearity of CNNs and the challenging non-linearity of plant growth appearance, care should be taken in extrapolating growth stage recognition results between classes. This is supported by the variable changes in performance between classes (Figure 10). On the other hand, if classes are designated by size, then there may be a more predictable relationship. This is unlikely, given a more effective approach would be to train a single-class model and split predictions by size. Further research should investigate the cross-dataset applicability of growth stage recognition performed here and if the results are species- and dataset-specific.

There was no clear indication that more recent versions (v6, v7, or v8) provided significant advantages for growth stage recognition over earlier iterations. As raised by Dang et al. (2023), this is likely the result of each development being specific to and focused on improving performance on the COCO dataset to optimize and fine-tune algorithms on those images. The accepted standard in the literature, by which algorithms are compared, is the COCO dataset; however, fine-tuning hyperparameters (such as level of image augmentation) and architecture is highly specific to a dataset. This result further emphasizes the importance of access to general yet standardized, open-source image datasets for the fine-tuning and development of plant recognition-specific architectures rather than simply transfer learning (Coleman and Salter, 2023; Danilevicz et al., 2021). For example, Güldenring & Nalpantidis (2021) found that pre-trained weights specific to agriculture provided benefits to the efficiency of training classification tasks. Given that the performance was similar among all the largest variants between YOLO versions, the differentiating factor is likely to be the improved ease of use, larger community engagement, ongoing development, and easier deployment. As it stands at the time of writing, YOLO v5 and v8 by Ultralytics improve the accessibility of the technology for non-machine learning specialists in plant science, with continuing updates adding more features. This trend is likely to continue with the incorporation of coding support tools, better documentation, and larger communities deploying models at the edge.

Besides accessibility, the improvements over time are more so in the efficiency of the network for real-time performance on embedded devices. It suggests that for complex problems such as growth stage detection, there needs to be a substantial focus on data. This should be two-fold. Firstly, data quantity is critical; however, there must also be attention to the classes used and the way in which those classes are identified. While not covered here, future investigations should seek alternative annotation strategies that target specific plant organs (flowers, leaves, stems) and then combine spatial context with detection information for a growth stage prediction. In this way, the continuum of plant growth stage can be reduced to discrete detection problems. The approach adopted in wheat head detection for high throughput phenotyping (David et al., 2020; Khaki et al., 2022) would be highly relevant. These studies present methods of detecting wheat ears specifically, which could be translated to *A. palmeri* inflorescences or other plant reproductive organs for flowering detection.

*4.3  Class activation maps (CAM)*

The use of CAM to visualize model attention is important in understanding the relationship between plants and how they are detected. It is rarely conducted in image-based precision agriculture, with examples including de Camargo et al. (2021), Güldenring & Nalpantidis (2021), Wang et al. (2020) and Espejo-Garcia et al. (2023); however, it presents an opportunity to ensure that algorithm performance is robust and less influenced by external biasing factors. It may also assist in targeting further data capture and algorithm training. Further research should focus on their use for understanding specific plant attributes that may contribute to effective classification of growth stage class.

## 5    Conclusion

We found that differentiating the highly variable plant growth stages of *A. palmeri* was possible with the latest versions of YOLO object detection models. In a dataset that combined both seedling and large, mature, flowering weeds, the smallest growth stages were the most likely to be missed at low resolutions, even with larger models, while larger growth stages were impacted by inter-class confusion, lowering overall performance. If small, seedling weeds are the intended recognition target, findings suggest the use of large resolutions with smaller model architectures. While the largest variants of each version had similarly high levels of recognition performance, YOLO v5 and v8 are supported by a large community with ongoing development and many tools for improving performance. Fortunately, some level of standardisation among the latest variants enables tools developed for one to be adapted readily for others. As versions and variants continually improve, it will be necessary to continue benchmarking their performance on difficult plant and plant phenology datasets. The detection of growth stages with off-the-shelf, open-source algorithms highlights the substantial opportunity for improving plant phenotyping and weed recognition technologies.

## 6    Conflict of Interest

The authors declare no conflict of interest.

## 7    Acknowledgements

Guy Coleman's visiting PhD research program at Texas A&M University was supported by an Australian-American Fulbright Commission scholarship. The research was also funded in part by Cotton Incorporated. The authors would like to thank Daniel Hathcoat and Daniel Lavy for maintaining the field trials and Sarah Kezar for advice with *A. palmeri* annotation.